%% file: main.tex
\documentclass{article}

\usepackage[nonatbib, preprint]{neurips_2022}

\usepackage[utf8]{inputenc} 
\usepackage[T1]{fontenc}    
\usepackage{hyperref}       
\usepackage{url}            
\usepackage{booktabs}       
\usepackage{amsfonts}       
\usepackage{nicefrac}       
\usepackage{microtype}      
\usepackage{xcolor}         

\usepackage{algorithm}
\usepackage{algorithmic}
\usepackage{enumitem}
\usepackage{amsmath,amssymb} 
\usepackage{xspace}
\usepackage{graphicx}
\usepackage{bbm}
\usepackage{multirow}
\usepackage{subcaption}

\usepackage{arydshln}

\newcommand{\tabincell}[2]{\begin{tabular}{@{}#1@{}}#2\end{tabular}}

\newlength\savewidth\newcommand\shline{\noalign{\global\savewidth\arrayrulewidth
  \global\arrayrulewidth 1pt}\hline\noalign{\global\arrayrulewidth\savewidth}}

\input{maths}

\newcommand{\methodshortname}{FocusFormer\xspace}

\makeatletter
\def\@fnsymbol#1{\ensuremath{\ifcase#1\or \dagger\or \ddagger\or
   \mathsection\or \mathparagraph\or \|\or **\or \dagger\dagger
   \or \ddagger\ddagger \else\@ctrerr\fi}}
\makeatother

\title{FocusFormer: Focusing on What We Need \\ via Architecture Sampler}

\author{%
  Jing Liu
  \quad Jianfei Cai
  \quad Bohan Zhuang\thanks{Corresponding author. Email: $\tt bohan.zhuang@monash.edu$} \\[0.2cm]
  Department of Data Science \& AI, Monash University, Australia
}

\begin{document}

\maketitle

\begin{abstract}
Vision Transformers (ViTs) have underpinned the recent breakthroughs in computer vision. However, designing the architectures of ViTs is laborious and heavily relies on expert knowledge. 
To automate the design process and incorporate deployment flexibility, one-shot neural architecture search decouples the supernet training and architecture specialization for diverse deployment scenarios. To cope with an enormous number of sub-networks in the supernet, existing methods treat all architectures equally important and randomly sample some of them in each update step during training. During architecture search, these methods focus on finding architectures on the Pareto frontier of performance and resource consumption, which forms a gap between training and deployment.
In this paper, we devise a simple yet effective method, called \methodshortname, to bridge such a gap. 
To this end, we propose to learn an architecture sampler to assign higher sampling probabilities to those architectures on the Pareto frontier under different resource constraints during supernet training, making them sufficiently optimized and hence improving their performance. During specialization, we can directly use the well-trained architecture sampler to obtain accurate architectures satisfying the given resource constraint, which significantly improves the search efficiency. 
Extensive experiments on CIFAR-100 and ImageNet show that our \methodshortname is able to improve the performance of the searched architectures while significantly reducing the search cost. For example, on ImageNet, our \methodshortname-Ti with 1.4G FLOPs outperforms AutoFormer-Ti by 0.5\% in terms of the Top-1 accuracy.
\end{abstract}

\section{Introduction}
\label{seq:intro}
\input{introduction}

\section{Related Work}
\input{related_work}

\section{\methodshortname}
\label{sec:method}
\input{method}

\section{Experiments}
\label{sec:experiments}
\input{experiments}

\section{Conclusion and Future Work}
\label{sec:conclusion}
In this paper, we have proposed \methodshortname to mitigate the gap between training and deployment in one-shot neural architecture search. To this end, we have devised an architecture sampler to obtain ViT architectures that are likely on the Pareto frontier under diverse resource constraints. By jointly training the architecture sampler with network parameters, we put more training resources on these salient architectures and hence improve the performance of the searched ViT architectures.
Once the supernet and architecture sampler have been well-trained, we can directly use our proposed architecture sampler 
to obtain architectures with promising performance while satisfying the given resource constraint. Experiments on CIFAR-100 and ImageNet have demonstrated that our proposed method is able to improve the performance of the searched architectures while significantly reducing the search cost. In the future, we may extend our method in several aspects. First, we may consider improving both the training and search efficiency by using some sparsity training strategies. Second, we may jointly perform neural architecture search and quantization simultaneously to obtain more compact ViTs with comparable performance.

\bibliographystyle{abbrv}
{
    \small
	\bibliography{reference}
}

\input{supp}

\end{document}

%% file: maths.tex
\def\st{\mbox{s.t.}}
\def\eg{\emph{e.g.}}
\def\ie{\emph{i.e.}}
\def\wrt{\mbox{w.r.t.}}
\def\etc{\emph{etc.}} 

\def\mA{{\mathcal A}}
\def\mB{{\mathcal B}}

\def\mD{{\mathcal D}}

\def\mL{{\mathcal L}}

\def\mN{{\mathcal N}}

\def\0{{\bf 0}}
\def\1{{\bf 1}}

\def\citep{\cite}
\def\citet{\cite}

\def\vs{\emph{vs.~}}

%% file: introduction.tex
With powerful computing resources and large amounts of labeled data, we have witnessed the tremendous successful applications of ViTs in computer vision~\cite{touvron2021training,dosovitskiy2021an,liu2021swin}. 
To push the state-of-the-art performance boundary, several studies~\cite{Wang_2021_ICCV,liu2021swin,chu2021twins,vaswani2017attention} have been proposed to craft the architectures of ViTs and achieved competitive performance in many visual tasks, such as image classification~\cite{dosovitskiy2021an,touvron2021training,yuan2021tokens} and dense prediction~\cite{carion2020end,zheng2021rethinking,cheng2021per}. However, manually designing ViT architectures is laborious and human expertise is often sub-optimal in exploring the huge design space.

To automatically design ViT architectures, much effort has been devoted to neural architecture search (NAS)~\cite{pham2018efficient,DARTS2019,cai2018proxylessnas,so2019evolved} by
finding
optimal architectures in a predefined search space.
To obtain compact architectures, some works~\cite{DARTS2019,wu2019fbnet,zhao2021memory} propose to search architectures for a specific resource constraint or hardware requirement. Nevertheless, when it comes to diverse platforms 
under various resource constraints (\eg, FLOPs, latency, on-chip memory), these methods need to design architecture for each scenario from scratch and the search cost grows linearly with the number of possible cases, which is extremely computationally expensive.
To tackle the above issue, one-shot NAS methods~\cite{guo2020single,Fan2020Reducing,xu2021autoformer,gong2022nasvit,chen2021searching} have been proposed to decouple the model training and the architecture search stage, where the first stage trains an over-parameterized supernet that covers a considerable number of sub-networks (\eg, $10^{20}$) with diverse architectural configurations and the second stage selects an optimal sub-network given the target resource constraint during deployment.

\begin{figure*}[t]
    \centering
    \begin{subfigure}{0.4\textwidth}
        \centering
        \includegraphics[height=2in]{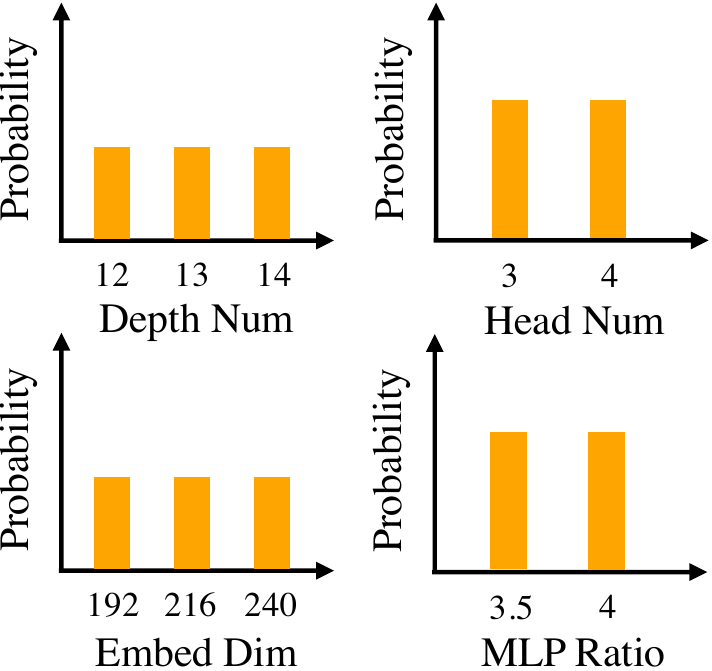}
        \caption{Uniform distribution of each dimension.}
        \label{fig:uniform_arch_dist}
    \end{subfigure}
    \hspace{0.02\textwidth}
    \begin{subfigure}{0.55\textwidth}
        \centering
        \includegraphics[height=2in]{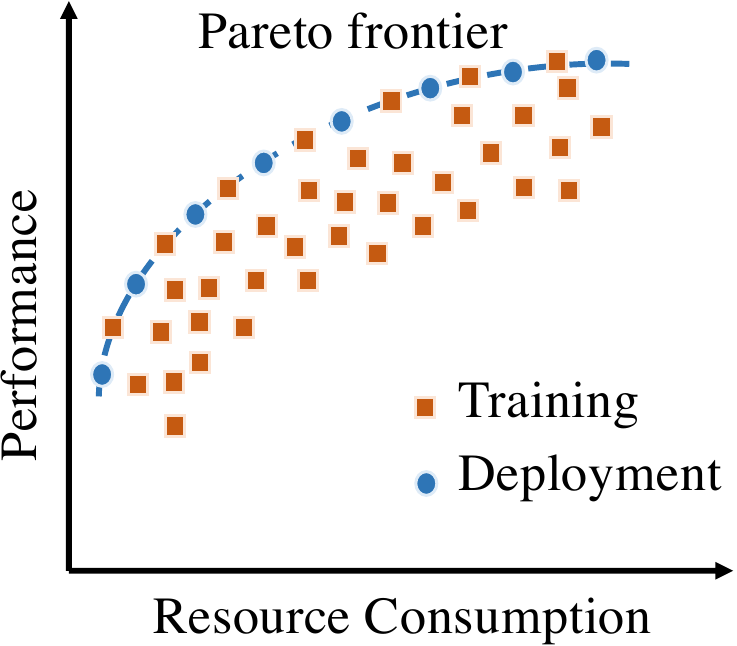}
        \caption{Gap between training and deployment in one-shot NAS.}
        \label{fig:gap_train_deployment}
    \end{subfigure}
    \caption{An overview of the uniform distribution of each architectural dimension and the gap between training and deployment in  one-shot NAS methods~\cite{xu2021autoformer,chen2021searching}. During supernet training, existing methods randomly sample sub-networks (rectangle) from a uniform architectural distribution in one update step.
    During deployment, these methods focus on the sub-networks (circle) on the Pareto frontier of performance and resource consumption. The other architectures that are not on the Pareto frontier are neglected.}
    \label{fig:gap_training_deployment}
    \vspace{-0.25in}
\end{figure*}

Nevertheless, there exists a gap between training and deployment. Specifically, to deal with the tremendous number of sub-networks in supernet, existing methods~\cite{guo2020single,Cai2020Once-for-All,wang2020hat,xu2021autoformer} randomly sample some of them from a uniform architectural distribution (See Figure~\ref{fig:uniform_arch_dist}) in each update step during training, where all architectures in the search space (rectangles in Figure~\ref{fig:gap_train_deployment}) have an equal probability of being chosen.
During deployment, these methods focus on those architectures (circles in Figure~\ref{fig:gap_train_deployment}) that are on the Pareto frontier of performance and resource consumption. The other architectures (rectangles not on the Pareto frontier) that have poor performance or do not meet the resource constraint are neglected. In this case, the training stage does not concentrate on the architectures on the Pareto frontier. As a result, the architectures with good performance may not be sufficiently trained and hence the performance of the searched architectures is sub-optimal.
Moreover, after supernet training, existing methods have to either use evolutionary search~\cite{guo2020single,Cai2020Once-for-All,xu2021autoformer,chen2021searching} or train a controller~\cite{xie2019exploring,ru2020neural} to search for optimal architectures given a deployment scenario, which requires extensive architecture evaluations and hence incurs high computational cost.

In this paper, we devise a simple yet effective method, called \emph{\methodshortname}, to bridge the gap between training and deployment. Unlike existing NAS methods for ViTs~\cite{xu2021autoformer,chen2021searching}, we assume and highlight that 
not all sub-networks in the supernet are equally important. 
Based on this intuition, we devise a parameterized architecture sampler to learn the architectural distribution under diverse resource constraints, 
where we consider four important dimensions of ViTs including depth, embedding dimension, multi-layer perceptron (MLP) ratio, and head number.
Instead of sampling from a uniform architectural distribution, we optimize the architecture sampler to generate sampling distributions conditioned on different resource constraints to assign higher probabilities to accurate sub-networks in an update step. 
In this way, we are able to train sub-networks that are likely to be on the Pareto frontier with more training budgets, thereby improving their performance.
Once the supernet and architecture sampler have been well-trained, we can directly use the architecture sampler to instantly generate candidate ViT architectures with good performance given the target resource constraint in the architecture search stage. Therefore, our \methodshortname does not need to involve extensive architecture evaluations such as in evolutionary methods~\cite{xu2021autoformer,real2019regularized} and 
hence greatly improve the search efficiency. Extensive experiments on CIFAR-100 and ImageNet show that our proposed method improves the performance of the searched architectures while significantly reducing the search cost. 

Our main contributions are summarized as follows: 
\begin{itemize}[leftmargin=*]
    \item 
    We propose \methodshortname to bridge the gap between training and deployment in one-shot NAS for ViTs. To our knowledge, this problem has not been well studied.  
    \item
We design a parameterized architecture sampler to serve as the bridge. During supernet training, the architecture sampler is jointly optimized with network parameters to identify and assign higher sampling probabilities for architectures on the Pareto frontier under different resource constraints, that align with the searched networks at the specialization stage. In this way, we allocate more training resources to these salient architectures and thus improve their performance. During deployment, we can instantly find the target candidate architectures for a given resource constraint using the well-trained architecture sampler, with very high search efficiency.
\item
    We evaluate our proposed method on CIFAR-100 and ImageNet. Extensive experiments demonstrate that our proposed method is able to improve the performance of the specialized network while greatly reducing the search cost. For example, on ImageNet, our \methodshortname-Ti surpasses AutoFormer-Ti by 0.5\% on the Top-1 accuracy.
\end{itemize}

%% file: related_work.tex
\noindent\textbf{Vision Transformers.} 
Recently, ViTs~\cite{dosovitskiy2021an,carion2020end,liu2021swin,zheng2021rethinking} have shown great representational power in computer vision. Specifically, a ViT~\cite{dosovitskiy2021an} contains a patch embedding layer, multiple Transformer blocks, and a task-dependent head, where each Transformer block consists of a multi-head self-attention (MSA) block and an MLP block. LayerNorm (LN)~\cite{ba2016layer} and residual connection~\cite{he2016deep} are applied before and after each block, respectively. To improve the performance of ViTs, several methods are proposed, including but not limited to incorporating hierarchical representations~\cite{fan2021multiscale,pan2021scalable,liu2021swin,Wang_2021_ICCV}, introducing inductive bias by inserting convolutional layers~\cite{wu2021cvt,li2021localvit,xu2021vitae,guo2021cmt,d2021convit,chen2021x,graham2021levit,he2021pruning} and improving positional encoding~\cite{chu2021conditional,wu2021rethinking}, \etc~Though achieving promising performance, the ViT architectures of these methods are manually designed, which may not fully explore the architecture space. Compared with these methods, our proposed \methodshortname focuses on automatically designing the ViT architecture.

\noindent\textbf{Neural architecture search.} Neural architecture search (NAS)~\cite{zoph2016neural,pham2018efficient,DARTS2019} seeks to automatically find architectures with promising performance while satisfying the resource constraint $B$, which can be formulated as
\begin{equation}
\begin{aligned}
    \label{eq:nas_problem}
    \min_{\alpha}&~\mL_{\mathrm{val}}(W_{\alpha}^*; \mD_{\mathrm{val}}) \\
    \st~ W_{\alpha}^* = \arg\min_{W_{\alpha}}&~\mL_{\mathrm{train}}(W_{\alpha}; \mD_{\mathrm{train}}),~C(\alpha) < B,
\end{aligned}  
\end{equation}
where $W_{\alpha}$ is the weights of the network with architecture $\alpha$, $C(\alpha)$ is the computational cost of $\alpha$, and $\mL_{\mathrm{train}}$, $\mL_{\mathrm{val}}$ are the training loss and validation loss, respectively. Here, $\mD_{\mathrm{train}}$ and $\mD_{\mathrm{val}}$ are the training and validation dataset, respectively. To solve Problem~(\ref{eq:nas_problem}), we have to train weights $W_{\alpha}$ for each architecture $\alpha$
until convergence to obtain $W_{\alpha}^*$, which takes unbearable computational cost. 
Most existing approaches either use random search~\cite{li2020random,bender2018understanding}, reinforcement learning~\cite{zoph2016neural,guo2020breaking,pham2018efficient,tan2019mnasnet}, evolutionary search~\cite{lu2019nsga,lu2020nsganetv2,real2019regularized,real2017large} or gradient-based optimization~\cite{DARTS2019,cai2018proxylessnas,wu2019fbnet,he2020milenas} to find optimal architectures.

In addition, one-shot NAS~\cite{Cai2020Once-for-All,guo2020single,Fan2020Reducing,xu2021autoformer,chen2021searching,gong2022nasvit} has been proposed to decouple the model training stage and architecture search stage. In the model training stage, the goal is to train an over-parameterized supernet that supports many sub-networks of different configurations with the shared parameters to accommodate different deployment scenarios. Let $W$ be the weights of the supernet. The objective function for the supernet training can be formulated as
\begin{equation}
    \label{eq:objective_one_shot}
     W^* = \arg\min_W \mathbb{E}_{\alpha \in \mA}\left[ \mL_{\mathrm{train}}(S(W, \alpha)) \right],
\end{equation}
where $S(W, \alpha)$ is a selection function that chooses a part of parameters from $W$ to constitute the sub-network with architecture configuration $\alpha$ and $\mA$ is the architecture search space. To solve Problem~(\ref{eq:objective_one_shot}), existing methods~\cite{xu2021autoformer,chen2021searching} assume that all the sub-networks are equally important and use a uniform sampling strategy to approximate the expectation term. Considering different resource constraints, Problem~(\ref{eq:objective_one_shot}) can be rewritten as
\begin{equation}
    \label{eq:objective_one_shot_rewritten}
     W^* = \arg\min_W \mathbb{E}_{B \sim \mB} \left[ \mathbb{E}_{\alpha \sim U(\cdot | B)}\left[ \mL_{\mathrm{train}}(S(W, \alpha)) \right] \right],
\end{equation}
where $B$ is a resource constraint sampled from a prior distribution $\mB$ and $U(\cdot | B)$ is a uniform distribution over architectures conditioned on $B$. 
For the architecture search stage, the goal is to obtain the specialized sub-network that satisfies the resource constraint $B$ given a deployment scenario while maximizing the validation accuracy $\mathrm{Acc}_{\mathrm{val}}(S(W^*, \alpha))$, which can be formulated as
\begin{equation}
    \begin{aligned}
    \label{eq:search_objective}
    \max_{\alpha}&~\mathrm{Acc}_{\mathrm{val}}(S(W^*, \alpha)) \\
    \st&~ C(\alpha) < B.
    \end{aligned}
\end{equation}
Here, $W^*$ is the well-learned weight of the supernet in the model training stage. Except for the validation accuracy, one can also use the negative validation loss to measure the performance of the searched architectures~\cite{DARTS2019,cai2018proxylessnas}. To solve Problem~(\ref{eq:search_objective}), existing methods either use evolutionary search~\cite{guo2020single,Cai2020Once-for-All,yu2020bignas,wang2020hat,xu2021autoformer,chen2021searching} or train a controller to search for optimal architectures~\cite{xie2019exploring,ru2020neural,huang2021searching}. Since the supernet has been well-trained, we are able to obtain any sub-network without any fine-tuning or retraining during the architecture search, which yields a much lower search cost compared with Problem~(\ref{eq:nas_problem}). Compared with AutoFormer~\cite{xu2021autoformer} and S3~\cite{chen2021searching}, our proposed \methodshortname bridges the gap between training and deployment, which greatly improves the performance of the specialized sub-networks and search efficiency. Compared with GreedyNAS~\cite{you2020greedynas,huang2021greedynasv2} and AttentiveNAS~\cite{wang2021attentivenas}, our \methodshortname differs in several aspects. First, our method focuses on searching ViT architectures while GreedyNAS and AttentiveNAS concentrate on searching the architectures of convolutional neural networks (CNNs). Second, they need to sample and rank multiple architectures for each iteration during supernet training, which introduces extensive computational overhead. In contrast, our \methodshortname trains an architecture sampler to obtain sub-networks in an efficient way. Third, they need to use evolutionary search to find architectures, which is computationally expensive. Our \methodshortname uses the well-trained architecture sampler to instantly generate ViT candidate architectures, which is extremely efficient.

%% file: method.tex
In one-shot NAS, the objectives for the model training stage in Eq.~(\ref{eq:objective_one_shot_rewritten}) and the architecture search stage in Eq.~(\ref{eq:search_objective}) are different, resulting in a gap between training and deployment. Specifically, existing methods~\cite{Cai2020Once-for-All,yu2020bignas} assume that all sub-networks are equally important and perform a uniform sampling strategy during supernet training. 
However, during deployment, we only care about architectures on the Pareto frontier of performance under various resource constraints.
In this case, the model training stage does not tailor for those specialized architectures on the best Pareto front.
As a result, the architectures with good performance may not be sufficiently trained and hence lead to the sub-optimal performance.
Moreover, once the supernet has been well-trained, we have to use either evolutionary search~\cite{guo2020single,Cai2020Once-for-All} or train a controller~\cite{xie2019exploring,ru2020neural} to search for optimal architectures due to the different objectives between the model training and architecture search stages, which introduces additional training cost.

\begin{figure*}[t]
    \centering
    \includegraphics[width=1.0\linewidth]{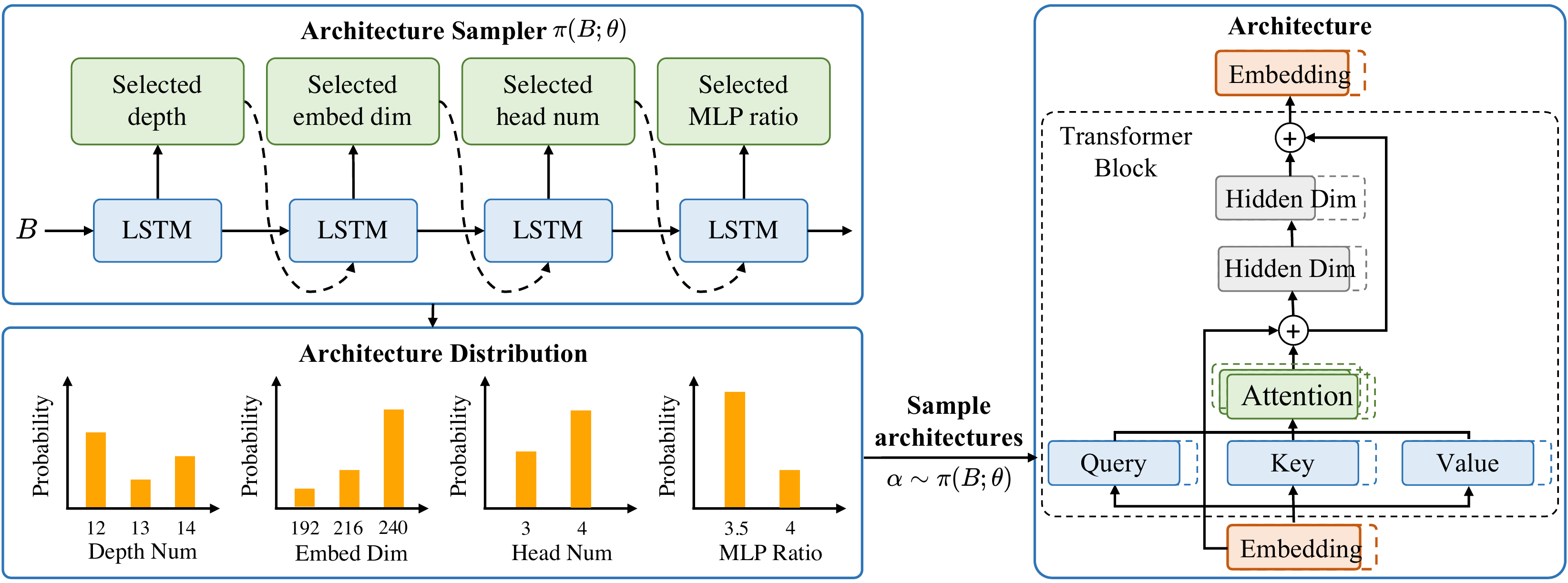}
    \caption{An illustration of our proposed architecture sampler. We formulate the architecture search problem as a sequential prediction problem where each element of the sequence denotes an architectural dimension of a ViT. Starting from a prior resource distribution $\mB$, we first sample a resource constraint $B\sim \mB$. We then feed $B$ into an LSTM network with four fully connected layers and output the architectural distribution $\pi(B;\theta)$. Last, we obtain the candidate architecture by performing sampling from the architectural distribution $\alpha \sim \pi(B;\theta)$.
    }
    \label{fig:architecture_sampler}
    \vspace{-0.2in}
\end{figure*}

In this paper, we jointly learn an architecture sampler with network parameters to assign higher probabilities to architectures that are likely on the Pareto frontier under different resource constraints during supernet training. In this way, we put more training resources on these important ViT architectures and hence mitigate the gap between training and deployment. Once the supernet and architecture sampler have been well-trained, we can directly use our architecture sampler to select good candidate architectures given a specific resource constraint, which is extremely efficient.

\subsection{Training Supernet with Architecture Sampler}
Given a ViT supernet $\mN$ with parameter $W$, we design an architecture sampler $\pi(\cdot;\theta)$ parameterized by $\theta$ to obtain architectures with potentially good performance given arbitrary $B$.
With the architecture sampler, we formulate the objective for the supernet training as
\begin{equation}
    \label{eq:objective_one_shot_sampler}
     W^* = \arg\min_W \mathbb{E}_{B \sim \mB} \left[ \mathbb{E}_{\alpha \sim \pi(B; \theta)}\left[ \mL_{\mathrm{train}}(S(W, \alpha)) \right] \right].
\end{equation}
Here, if we replace $\pi(B;\theta)$ with the uniform distribution over architectures $U(\cdot | B)$ conditioned on  $B$, Problem~(\ref{eq:objective_one_shot_sampler}) is reduced to the original one-shot NAS problem in Eq.~(\ref{eq:objective_one_shot_rewritten}). Equipped with $\pi(\cdot;\theta)$, those ViT architectures that are potentially on the Pareto frontier are trained with more computing resources.
In this way, the gap between training and deployment can be mitigated and hence improve the performance of the searched architectures under different resource constraints. We summarize our training pipeline in Algorithm~\ref{alg:supernet_training}.
Note that training the architecture sampler for each epoch is computationally expensive and not necessary. To reduce the computational cost, we only update our architecture sampler per $\tau$ epoch.

\begin{algorithm}[t]
    \caption{Training algorithm for FocusFormer.}
	\begin{algorithmic}[1]\small
	\REQUIRE Supernet $\mN$ with parameter $W$, architecture sampler $\pi(\cdot; \theta)$ with parameter $\theta$, the number of training epoch $T$, training dataset $\mD_{\mathrm{train}}$, architecture sampler update interval $\tau$, resource constraint distribution $\mB$, learning rate $\eta$, hyper-parameter $\beta$.
        \STATE Randomly initialize the supernet parameter $W$ and architecture sampler parameter $\theta$.
    	\FOR{$t \in \{1, 2,\dots,T\}$}
    	    \IF{$t = n\tau, ~~n \in \mathbb{N}^{+}$}
    	        \STATE Train architecture sampler $\pi(B;\theta)$ on $\mD_{\mathrm{train}}$.
    	    \ENDIF
    	    \FOR{each batch of data in $\mD_{\mathrm{train}}$}
    	    \STATE Sample a resource constraint $B \sim \mB$.
    	    \STATE Sample an architecture $\alpha \sim \pi(B; \theta)$.
    	    \STATE Train supernet $\mN$ with $\alpha$.
    	    \ENDFOR
    	\ENDFOR
	\end{algorithmic}
	\label{alg:supernet_training}
\end{algorithm}

\subsection{Learning Architecture Sampler}
\label{sec:learn_sampler}
To determine the ViT architecture, we formulate the NAS problem as a sequence prediction problem where each element of a sequence denotes an architectural dimension (\eg, the depth of ViT) following~\cite{pham2018efficient,zoph2016neural}. To predict the architectural distribution, our architecture sampler $\pi(\cdot; \theta)$ is an LSTM network with four fully connected layers, which takes a resource constraint $B$ as input and then outputs four architectural dimensions of a ViT, including depth, embedding dimension, head number and MLP ratio, as shown in Figure~\ref{fig:architecture_sampler}. 
We then perform sampling from the architectural distribution to obtain architectures, \ie, $\alpha \sim \pi(B; \theta)$. To represent different $B$, we build learnable embedding vector for each $B$ following~\cite{pham2018efficient,guo2021pareto}.

To train $\pi(\cdot; \theta)$, we formulate the objective function as
\begin{equation}
    \label{eq:objective_sampler}
    \max_{\theta} \mathbb{E}_{B \sim \mB} \left[ \mathbb{E}_{\alpha \sim \pi(B; \theta)} \left[  R(W, \alpha) \right]  \right],
\end{equation}
where $R(W, \alpha)$ is a performance metric of the sampled architecture $\alpha$. To guide the training of $\pi(\cdot; \theta)$, we hope the resource consumption of $\alpha$ not only close to the given resource constraint $B$, but also achieves high accuracy following~\cite{bender2020can}. Considering both the resource constraint and accuracy, we have 
\begin{equation}
    \label{eq:reward_function}
    R(W, \alpha) = \mathrm{Acc}_{\mathrm{val}}(S(W,\alpha)) - \beta \left| \frac{B}{C(\alpha)} - 1 \right|,
\end{equation}
where $\mathrm{Acc_{\mathrm{val}}(S(W, \alpha)))}$ is the validation accuracy of $\alpha$ with corresponding weights $S(W, \alpha)$ and $\beta$ is a hyper-parameter that makes a balance between accuracy and resource consumption. In order to reduce the training cost of $\pi(B; \theta)$, we only compute the validation accuracy on a mini-batch of data rather than the whole validation data set. In general, Eq.~(\ref{eq:reward_function}) is not differentiable \wrt~ $\theta$. 
Following~\cite{zoph2016neural,pham2018efficient}, we use reinforcement learning with policy gradient~\cite{williams1992simple} to solve Problem~(\ref{eq:objective_sampler}). Specifically, we update $\theta$ by ascending the policy gradient, which can be formulated as:
\begin{equation}
    \label{eq:update_theta}
    \theta \leftarrow \theta + \eta R(W, \alpha) \nabla_{\theta} \log \pi(B; \theta),
\end{equation}
where $\eta$ is the learning rate of $\pi(\cdot;\theta)$. We summarized the training details of our proposed architecture sampler in Algorithm~\ref{alg:train_sampler}.

Note that our architecture sampler introduces some additional costs during supernet training.
To reduce the training cost, 
we train the supernet using a progressive learning strategy following~\cite{tan2021efficientnetv2,li2022autoprog}. Specifically, we gradually increase the image size from a small $S_0 \times S_0$ to the final $S_e \times S_e$ with a fixed patch size $p$. As a result, the number of patches is increased from $S_0^2 / p^2$ to $S_e^2 / p^2$. However, in ViT, the size of positional encodings is related to the number of patches. To overcome this limitation, we proposed to use conditional positional encodings following~\cite{chu2021conditional}.

\noindent\textbf{Resource constraint distribution.} In Eqs.~(\ref{eq:objective_one_shot_sampler}) and~(\ref{eq:objective_sampler}), we introduce a prior distribution $\mB$. For simplicity, we assume that $\mB$ is a uniform distribution, which indicates that all resource constraints have an equal probability of being sampled. Note that there are a large number of $B$ in $\mB$ and we can not enumerate them all during training. To overcome this limitation, we perform quantization to obtain discrete resource constraint as $B=\mathrm{round}(B/s) * s$, where $\mathrm{round}(x)$ returns the nearest integer of a given value $x$ and $s$ is a step size of the round function. To represent a resource constraint with an arbitrary value, we use the linear interpolation method following~\cite{guo2021pareto,radford2015unsupervised}.

\subsection{Efficient Architecture Search}
Once we have well trained the supernet and the proposed architecture sampler, we now focus on obtaining architectures that satisfy the given resource constraint while achieving promising performance. Using the well-learned $\pi(\cdot; \theta)$, we are able to directly obtain good candidate architectures.
Specifically, given a resource constraint $B$, we first sample multiple architectures from the predicted architectural distribution $\pi(B;\theta)$. We will repeat the sampling process if the sampled ViT architecture does not satisfy the resource constraint, \ie, $C(\alpha) > B$. Last, we select the ViT architecture with the highest validation accuracy as the searched architecture. Since only a limited number of architecture evaluations are involved, the search cost of our method is much smaller than the existing one-shot NAS methods~\cite{li2022autoprog,chen2021searching}.

\begin{algorithm}[t]
    \caption{Training algorithm for architecture sampler.}
    \begin{algorithmic}[1]\small
	\REQUIRE Supernet $\mN$, architecture sampler $\pi(\cdot; \theta)$ with parameter $\theta$, total training iterations $K$, training dataset $\mD_{\mathrm{train}}$, distribution of resource constraint $\mB$, learning rate $\eta$, hyper-parameter $\beta$.
    \FOR{$k \in \{1, 2,\dots,K\}$}
        \STATE Sample a batch of data from $\mD_{\mathrm{train}}$.
        \STATE Sample a resource constraint $B \sim \mB$.
        \STATE Sample an architecture $\alpha \sim \pi(B; \theta)$.
        \STATE Compute the reward $R(W, \alpha)$ using Eq.~(\ref{eq:reward_function}).
        \STATE Update $\theta$ using Eq.~(\ref{eq:update_theta}).
    \ENDFOR
    \end{algorithmic}
    \label{alg:train_sampler}
\end{algorithm}

%% file: experiments.tex
\noindent\textbf{Datasets.} We evaluate the proposed method on two image classification datasets, namely, CIFAR-100~\cite{Krizhevsky2009LearningML} and ImageNet~\cite{deng2009imagenet}. CIFAR-100 contains 100 classes, where each class has 500 training images and 100 testing images. ImageNet is a large-scale dataset that contains 1.28M training images and 50k validation images with 1k classes. To construct the validation set, we randomly sample 10k and 100k training samples from CIFAR-10 and ImageNet following~\cite{DARTS2019,xu2021autoformer}.

\noindent\textbf{Search space.} 
Following~\cite{xu2021autoformer}, we apply our proposed method to ViT search space with three different settings, namely, supernet-tiny, supernet-small and supernet-base. Each search space contains four main dimensions to build transformer blocks: embedding dimension, the number of heads, MLP ratio and network depth. The detailed settings of each search space can be found in AutoFormer~\cite{xu2021autoformer}. For convenience, we use ``\methodshortname-Ti'', ``\methodshortname-S'' and ``\methodshortname-B'' to represent the architectures obtained by \methodshortname in the tiny, small and base search space, respectively.

\begin{table*}[!tb]
\caption{
Performance comparisons of different methods on ImageNet. $^{*}$ denotes that we obtain the results from~\cite{touvron2021training}. $^{**}$ indicates that we change the model to use conditional positional encodings~\cite{chu2021conditional} for fair comparisons.
}
\centering
\renewcommand\arraystretch{1.2}
\scalebox{0.84}
{
\begin{tabular}{c|cc|ccc}
Models & Top-1 Acc. (\%) & Top-5 Acc. (\%) & \#Params (M) & FLOPs (G) & Resolution \\
\shline
ResNet-18~\cite{he2016deep} & 69.8& - & 11.7 & 1.8 & 224 \\
DeiT-Ti~\cite{touvron2021training} & 72.2 & 91.1 & 5.7 & 1.2 & 224 \\
AutoFormer-Ti$^{**}$~\cite{xu2021autoformer} & 74.6 & 92.3 & 5.7 & 1.3 & 224 \\
FocusFormer-Ti (Ours) & \textbf{75.1} & \textbf{92.8} & 6.2 & 1.4 & 224 \\
\shline
ResNet-50~\cite{he2016deep} & 76.1 & - & 25.6 & 4.1 & 224 \\
RegNetY-4GF$^{*}$~\cite{radosavovic2020designing} & 80.0 & - & 21.4 & 4.0 & 224 \\
BoTNet-S1-59~\cite{srinivas2021bottleneck} & 81.7 & 95.8 & 33.5 & 7.3 & 224 \\
T2T-ViT-14~\cite{yuan2021tokens} & 81.7 & - & 21.5 & 6.1 & 224 \\
DeiT-S~\cite{touvron2021training} & 79.9 & 95.0 & 22.1 & 4.7 & 224 \\
ViT-S/16~\cite{dosovitskiy2021an} & 78.8 & - & 22.1 & 4.7 & 384 \\
TNT-S~\cite{han2021transformer} & 81.5 & 95.7 & 23.8 & 5.2 & 224 \\ 
AutoFormer-S$^{**}$~\cite{xu2021autoformer} & 81.4 & 95.6 & 24.3 & 5.1 & 224 \\
FocusFormer-S (Ours) & \textbf{81.6} & \textbf{95.6} & 23.7 & 5.0 & 224 \\
\hdashline
ResNet152~\cite{he2016deep} & 78.3 & - & 60.2 & 11.6 & 224 \\
ResNeXt101-64x4d~\cite{xie2017aggregated} & 79.6 & - & 83.5 & 15.6 & 224 \\
ViT-B/16~\cite{dosovitskiy2021an} & 79.7 & - & 86.6 & 17.6 & 384 \\
DeiT-B~\cite{touvron2021training} & 81.8 & 95.6 & 86.6 & 17.6 & 224 \\
AutoFormer-B$^{**}$~\cite{xu2021autoformer} & 81.7 & 95.5 & 52.8 & 11.0 & 224 \\
FocusFormer-B (Ours) & \textbf{81.9} & \textbf{95.6} & 52.7 & 11.0 & 224 \\
\end{tabular}
}
\label{table:results_on_imagenet}
\vspace{-0.2in}
\end{table*}

\noindent\textbf{Evaluation metrics.}
We measure the performance of different methods using the Top-1 and Top-5 accuracy. Following~\cite{sandler2018mobilenetv2,howard2019searching}, we use the floating-point operations (FLOPs) and the number of parameters (\#Params) to measure the resource consumption and model size, respectively. Following~\cite{DARTS2019}, we also measure the training and search costs using the training and search durations on a single NVIDIA GeForce RTX 3090 GPU, respectively.

\noindent\textbf{Implementation details.} We train our model using a server with 8 V100 GPUs. Following~\cite{xu2021autoformer}, we train the supernet with the weight entanglement strategy. The supernet is trained for 500 epochs with a mini-batch size of 128 and 1024 on CIFAR-100 and ImageNet, respectively. We scale the learning rate according to the batch size with formula: $\mathrm{lr}_{\mathrm{scaled}}=\frac{5 \times 10^{-4}}{512} \times \mathrm{batchsize}$ and use cosine rule~\cite{loshchilov2016sgdr} to decrease the learning rate. We use AdamW~\cite{loshchilov2018decoupled} for optimization. Following DeiT~\cite{touvron2021training}, we use RandAugment~\cite{cubuk2020randaugment}, Cutmix~\cite{yun2019cutmix}, Mixup~\cite{zhang2018mixup} and random erasing~\cite{zhong2020random} except the repeated augmentation~\cite{hoffer2019augment} for data augmentation. For the progressive learning strategy, we set $S_e$ to 224. $S_0$ is set to 128 for experiments on CIFAR-100 and 160 for experiments on ImageNet. The patch size p is set to 16 following~\cite{dosovitskiy2021an}.
For the architecture sampler, we use the same batch size and optimization method as the supernet. 
We initialize the learning rate to $2.5 \times 10^{-4}$ and $1 \times 10^{-3}$ for the experiments on CIFAR-100 and ImageNet, respectively. We set the dimension of the hidden state for the LSTM network to 64. The number of training iterations $K$ for $\pi(\cdot;\theta)$ and the hyper-parameter $\beta$ in Eq.~(\ref{eq:reward_function}) are set to 100 and 0.07, respectively.
We set $\tau$ to 30 and 40 for the experiments on CIFAR-100 and ImageNet, respectively. 
For the architecture search stage, we randomly sample 30 architectures and select the architecture with the highest validation accuracy. All the searched architectures directly inherit weights from the learned supernet without any fine-tuning. 

\subsection{Main Results}
We apply our \methodshortname to search ViT architectures under different FLOPs on ImageNet. We show the results in Table~\ref{table:results_on_imagenet}. For fair comparisons, we modify AutoFormer to use conditional positional encodings~\cite{chu2021conditional}. From the results, our searched ViTs perform better than the CNN counterparts, such as ResNet~\cite{he2016deep}, RegNetY~\cite{radosavovic2020designing} and ResNeXt~\cite{xie2017aggregated}, which shows the strong representational power of ViTs. We can also observe that our proposed \methodshortname achieves much higher performance than the manually designed ViTs. For example, \methodshortname-Ti with 1.4G FLOPs outperforms DeiT-Ti~\cite{touvron2021training} by 2.9\% on the Top-1 accuracy. More critically, our \methodshortname consistently outperforms Autoformer~\cite{xu2021autoformer} under different resource constraints. For example, with 1.4G FLOPs, \methodshortname-Ti achieves 75.1\% in terms of the Top-1 accuracy, which is 0.5\% higher than AutoFormer-Ti.

\begin{figure}[!t]
    \centering
        \begin{minipage}[t]{0.47\textwidth}
        \centering
        \includegraphics[height=1.3in]{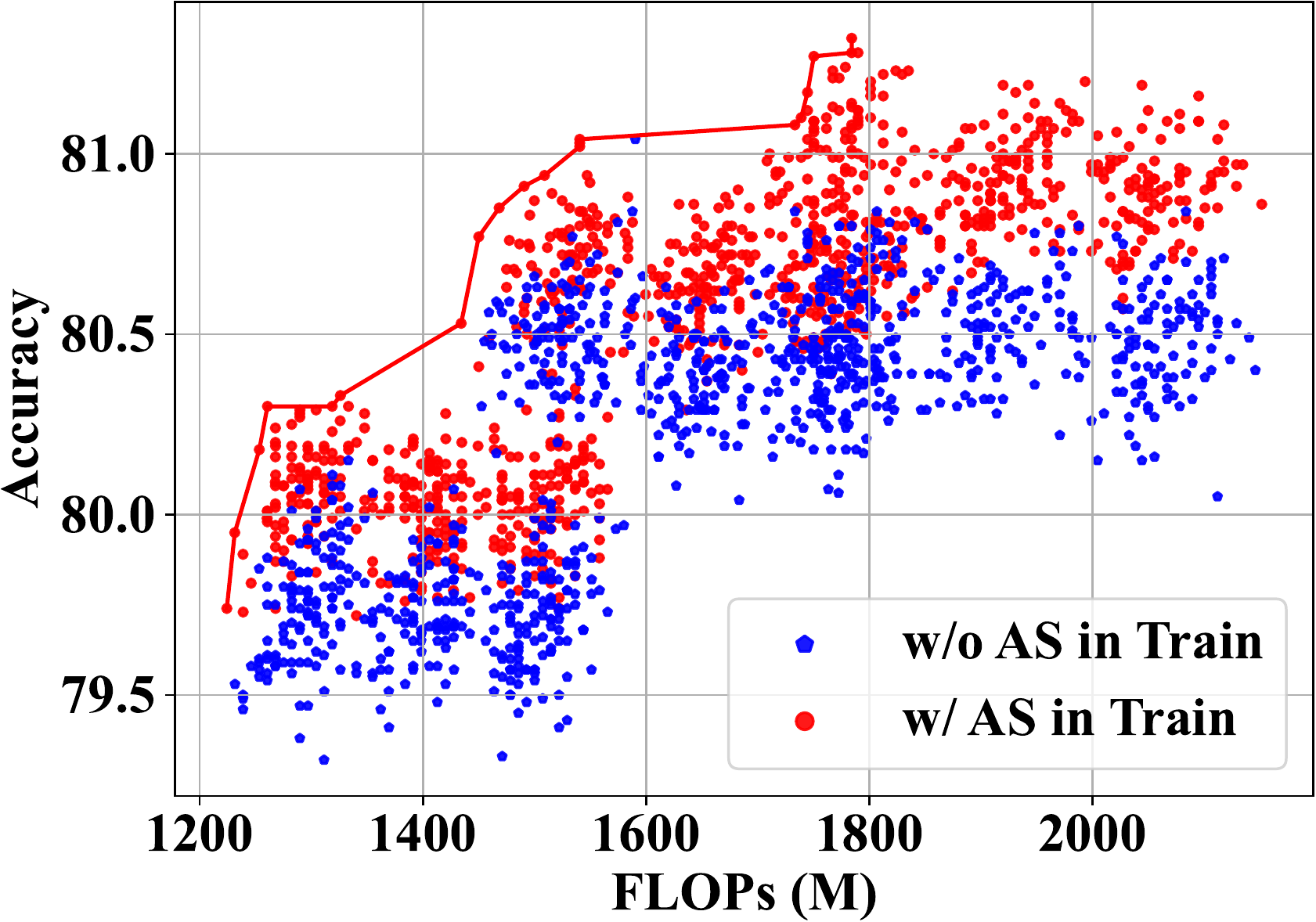}
        \caption{
        Top-1 accuracy of 1k sub-networks extracted from the supernet-tiny obtained with different training methods on CIFAR-100. 
        }
        \label{fig:effect_arch_sampler}
        \end{minipage}
        \hspace{0.1in}
        \begin{minipage}[t]{0.47\textwidth}
        \centering
        \includegraphics[height=1.3in]{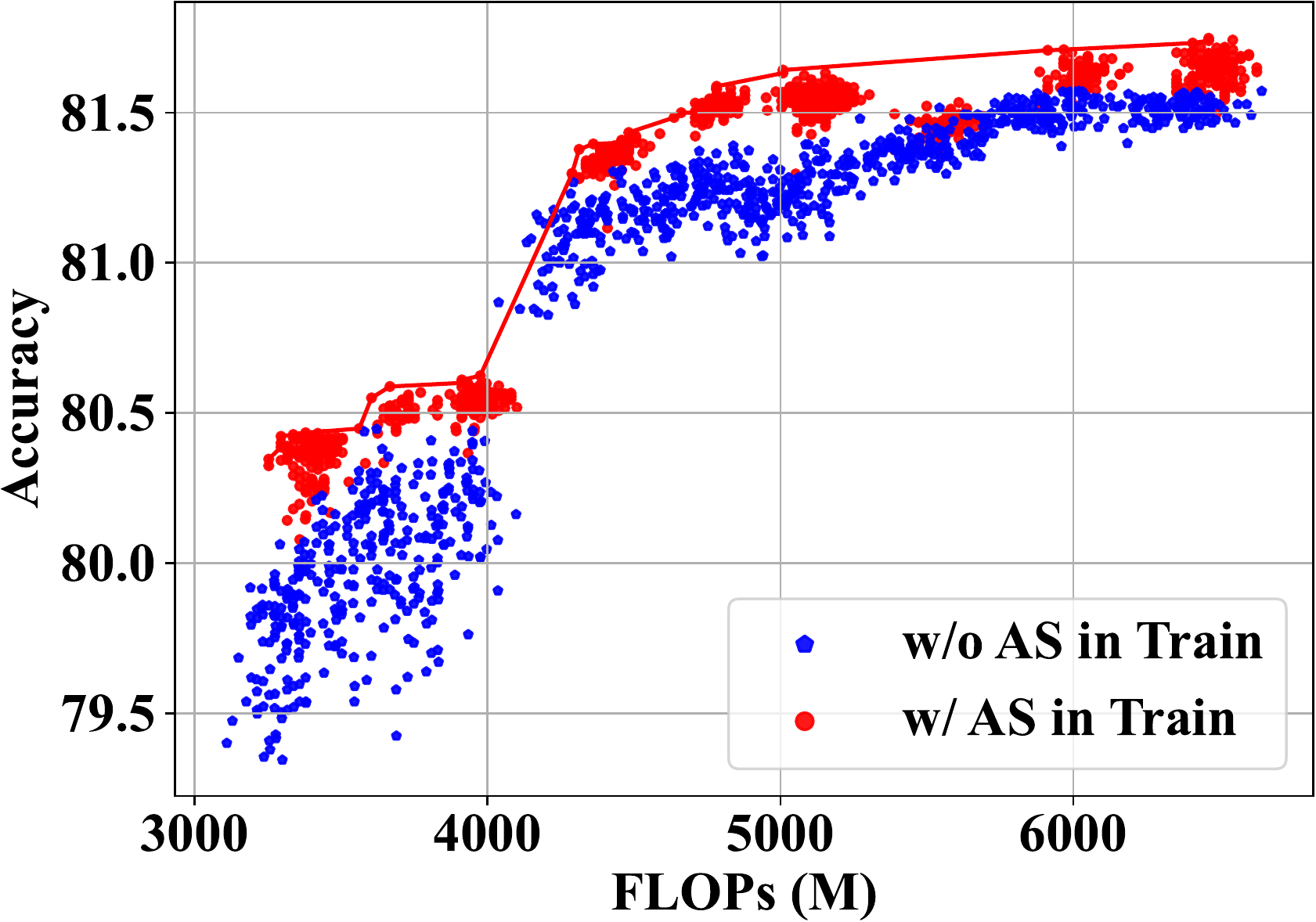}
        \caption{Top-1 accuracy of 1k sub-networks extracted from the supernet-small obtained with different training methods on ImageNet. }
        \label{fig:effect_arch_sampler_imagenet}
        \end{minipage}
        \vspace{-0.1in}
\end{figure}

\begin{table}[!tb]
\caption{
Effect of the architecture sampler. ``AS in Train'' denotes that we use architecture sampler rather than uniform sampling~\cite{xu2021autoformer} to obtain sub-networks during supernet training. ``AS in Search'' indicates that we use architecture sampler instead of evolutionary search~\cite{xu2021autoformer} to obtain candidate architectures during search. We report the results of FocusFormer-Ti obtained by different methods on CIFAR-100. 
``$^\dagger$'' denotes that we include the training cost of the architecture sampler.
}
\centering
\renewcommand\arraystretch{1.2}
\scalebox{0.75}
{
\begin{tabular}{cc|c|ccc}
\tabincell{c}{AS \\in Train} & \tabincell{c}{AS 
\\in Search} & \tabincell{c}{Top-1 \\Acc. (\%)} & \tabincell{c}{FLOPs (G)} & \tabincell{c}{Train Cost \\(GPU Hours)} & \tabincell{c}{Search Cost \\ (GPU Hours)} \\
\shline
& & 80.5 & 1.8 & 7.7 &1.7 \\
\checkmark & & \textbf{81.0} & 1.8 & 9.1 & 1.7 \\
& \checkmark & 80.7 & 1.7 & 7.7 & 0.3$^{\dagger}$ \\
\checkmark & \checkmark & \textbf{81.0} & 1.7 & 9.1 & $<0.1$ \\
\end{tabular}
}
\label{table:effect_arch_sampler}
\vspace{-0.1in}
\end{table}

\begin{table}[!tb]
\caption{
Effect of the progressive learning. ``PL'' represents that we use the progressive learning strategy~\cite{tan2021efficientnetv2} during supernet training. We report the results of FocusFormer-Ti obtained by different methods on CIFAR-100. 
}
\centering
\renewcommand\arraystretch{1.2}
\scalebox{0.75}
{
\begin{tabular}{c|cc|ccc}
Method & \tabincell{c}{Top-1 \\Acc. (\%)} & \tabincell{c}{Top-5 \\Acc. (\%)} & \tabincell{c}{FLOPs (G)} & \tabincell{c}{Train Cost \\(GPU Hours)} \\
\shline
\methodshortname-Ti w/o PL & 80.6 & 95.8 & 1.7 & 11.6 \\
\methodshortname-Ti w/ PL & \textbf{81.0} & \textbf{96.0} & 1.7 & 9.1 \\
\end{tabular}
}
\label{table:effect_progressive_learning}
\vspace{-0.1in}
\end{table}

\subsection{Further Studies}
\noindent\textbf{Effect of the architecture sampler.}
To investigate the effect of our architecture sampler (``AS''), we apply ``AS'' to different stages. Specifically, applying the ``AS'' in the supernet training stage denotes that we sample sub-networks from the learned distribution using ``AS'' instead of a uniform distribution~\cite{xu2021autoformer} while applying the ``AS'' in the search stage represents that we obtain candidate architectures using ``AS'' rather than the evolutionary methods~\cite{xu2021autoformer}. The results are provided in Table~\ref{table:effect_arch_sampler}. To show the supernet performance under various resource constraints, we also randomly sample 1k architectures from the supernet and show the results in Figures~\ref{fig:effect_arch_sampler} and~\ref{fig:effect_arch_sampler_imagenet}. 

From the results, using the ``AS'' in the supernet training stage introduces a little training cost but consistently improves the performance under various resource constraints (See Figures~\ref{fig:effect_arch_sampler} and~\ref{fig:effect_arch_sampler_imagenet}). The performance improvement is more significant in high FLOPs scenarios. For example, with about 1.7G FLOPs, our method with ``AS'' in the supernet training stage outperforms those without ``AS'' by 0.5\% in terms of the Top-1 accuracy on CIFAR-100. Moreover, using our ``AS'' in the search stage achieves comparable or even better performance than the evolutionary search while significantly reducing the search cost (0.3 \vs 1.7 GPU Hours). Using ``AS'' in both the training stage and search stage, our \methodshortname yields the best performance with 81.0\% Top-1 accuracy and 1.7G FLOPs while the search cost is less than 0.1 GPU Hour. These results show that our ``AS'' is able to improve the performance of the searched architectures while significantly reducing the search cost.

\noindent\textbf{Effect of the progressive learning.}
To investigate the effect of progressive learning (See Section~\ref{sec:learn_sampler}), we apply our method to obtain architectures with and without progressive learning during supernet training. From Table~\ref{table:effect_progressive_learning}, \methodshortname with ``PL'' strategy improves the supernet performance by 0.4\% on the Top-1 accuracy while significantly reducing the training cost by 2.5 GPU Hours. Therefore, we use progressive learning by default.

\begin{table}[!tb]
\caption{Performance comparisons with different update intervals $\tau$ on CIFAR-100.
}
\centering
\renewcommand\arraystretch{1.2}
\scalebox{0.75}
{
\begin{tabular}{c|c|cc|cc}
Model & $\tau$ & \tabincell{c}{Top-1 \\Acc. (\%)} & \tabincell{c}{Top-5 \\Acc. (\%)} & \tabincell{c}{\#Params (M)} & \tabincell{c}{FLOPs (G)} \\
\shline
\multirow{4}{*}{\methodshortname-Ti} & 10 & 80.3 & 95.9 & 8.1 & 1.8 \\
& 20 & \textbf{81.0} & \textbf{96.0} & 8.0 & 1.7 \\
& 30 & 80.8 & 96.1 & 7.8 & 1.7 \\
& 40 & 80.5 & 95.9 & 8.0 & 1.8 \\
\end{tabular}
}
\label{table:different_tau}
\vspace{-0.15in}
\end{table}

\noindent\textbf{Effect of different update intervals $\tau$.} To investigate the effect of different update intervals $\tau$ in the supernet training stage, we train \methodshortname-Ti with different $\tau \in \{10, 20, 30, 40\}$ and show the results in Table~\ref{table:different_tau}. From the results, with the decrease of $\tau$, the performance of the searched architecture becomes better first and then goes worse. For example, the searched architecture obtained by $\tau=30$ with 0.1G lower FLOPs surpasses that of $\tau=40$ by 0.3\% in terms of the Top-1 accuracy. One possible reason is that we need to update the architecture sampler regularly to a certain extent to accommodate the performance change during supernet training. However, too small $\tau$ indicates the frequent changes of the learned architectural distribution. As a result, the potentially good architectures may not be sufficiently trained.
Moreover, as our \methodshortname achieves the best performance when $\tau$ is set to 20, we use it by default in our experiments.

\noindent\textbf{Effect of different step sizes $s$.}
To investigate the effect of different step sizes $s$ in resource constraint discretization, we train \methodshortname-Ti with different $s \in \{ 50, 100, 200, 300, 400, 500 \}$. Here, smaller $s$ denotes more resource constraint candidates.
From Table~\ref{table:different_step_size},
the performance of searched architecture becomes better and then worsens with the decrease of $s$. On one hand, larger $s$ results in a small number of resource constraint candidates, which can not accurately represent an arbitrary resource constraint due to coarse-grained linear interpolation. For example, our method yields the worst performance with $s=500$.
 
On the other hand, smaller $s$ leads to a large number of resource constraint candidates, which introduces too many learnable embedding vectors and hence incurs the optimization difficulty.
With $s=200$,
the searched architectures achieve the best performance (81.0\% in terms of the Top-1 accuracy).

\begin{table}[!tb]
\caption{Performance comparisons with different step sizes $s$ on CIFAR-100.
}
\centering
\renewcommand\arraystretch{1.2}
\scalebox{0.75}
{
\begin{tabular}{c|c|cc|cc}
Model & $s$ & \tabincell{c}{Top-1 \\Acc. (\%)} & \tabincell{c}{Top-5 \\Acc. (\%)} & \tabincell{c}{\#Params (M)} & \tabincell{c}{FLOPs (G)} \\
\shline
\multirow{6}{*}{\methodshortname-Ti} & 50 & 80.4 & 96.0 & 7.7 & 1.7\\ 
& 100 & 80.4 & 95.8 & 7.9 & 1.7 \\
& 200 & \textbf{81.0} & \textbf{96.0} & 8.0 & 1.7 \\
& 300 & 80.5 & 95.8 & 7.9 & 1.7 \\
& 400 & 80.8 & 95.9 & 7.9 & 1.7 \\
& 500 & 80.3 & 95.7 & 7.7 & 1.7 \\
\end{tabular}
}
\label{table:different_step_size}
\vspace{-0.2in}
\end{table}

%% file: supp.tex
\newpage
\appendix

\renewcommand\thesection{\Alph{section}}
\renewcommand\thefigure{\Alph{figure}}
\renewcommand\thetable{\Alph{table}}
\renewcommand{\theequation}{\Alph{equation}}

\setcounter{table}{0}
\setcounter{figure}{0}

\begin{center}
    {
        \Large{\textbf{Supplementary Material for FocusFormer: Focusing on What We Need via Architecture Sampler}}
    }
\end{center}

\section{Visualization of the searched architectures}
\label{sec:visualization}
\begin{figure*}[!b]
    \centering
    \includegraphics[width=0.55\linewidth]{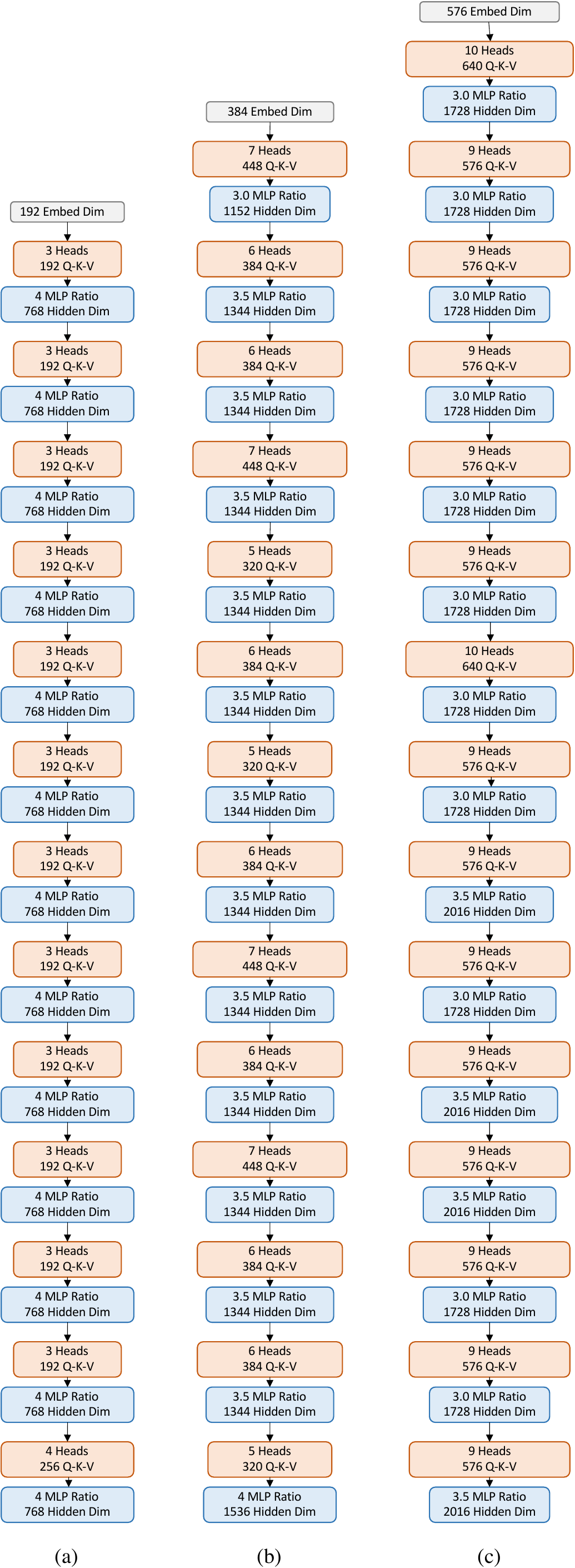}
    \caption{The architectures obtained by \methodshortname under different resource constraints on ImageNet. The orange and blue blocks denote the multi-head self-attention layers and MLP blocks, respectively. We omit the shortcuts and conditional positional encodings for simplicity. (a): The searched architecture of \methodshortname-Ti with 1.4G FLOPs and 6.2M \#Params. (b): The searched architecture of \methodshortname-S with 5.0G FLOPs and 23.7M \#Params. (c): The searched architecture of \methodshortname-B with 11.0G FLOPs and 52.7M \#Params.
    }
    \label{fig:visualization_arch}
\end{figure*}

In this section, we show the ViT architectures obtained by our \methodshortname in Figure~\ref{fig:visualization_arch}. The searched architectures tend to select large Q-K-V dimensions, MLP ratios, head numbers, and depth numbers with the increase of FLOPs and \#Params. For \methodshortname-S and \methodshortname-B, we observe that the intermediate Transformer blocks prefer large head numbers, while the deep blocks tend to choose small head numbers.

